\documentclass[11pt]{article}

\usepackage[final]{acl}

\usepackage{times}
\usepackage{latexsym}

\usepackage[T1]{fontenc}

\usepackage[utf8]{inputenc}

\usepackage{microtype}

\usepackage{inconsolata}

\usepackage{graphicx}
\usepackage{multirow}
\usepackage{booktabs}
\usepackage{siunitx}
\usepackage{amsmath}
\usepackage{arydshln}
\usepackage{makecell}

\title{Long Story Short: Disentangling Compositionality and Long-Caption Understanding in Contrastive VLMs}

\author{
  Israfel Salazar$^{1}$ \quad
  Desmond Elliott$^{1}$ \quad
  Yova Kementchedjhieva$^{2}$ \\
  $^{1}$Department of Computer Science, University of Copenhagen
  $^{2}$MBZUAI \quad \\
  \texttt{\{israfel.salazar, de\}@di.ku.dk} \\
  \texttt{yova.kementchedjhieva@mbzuai.ac.ae} \\
}

\begin{document}
\maketitle

\begin{abstract}
Contrastive vision-language models (VLMs) have made significant progress in binding visual and textual information, yet understanding long, compositional captions remains an open challenge. While these capabilities are often assumed to be closely related, the conditions under which they reinforce each other remain unclear. In this paper, we empirically analyze when compositional reasoning and long-caption understanding transfer across tasks, and when this relationship fails. Through controlled experiments across diverse training objectives, datasets, and architectural designs, we find a bidirectional but sensitive relationship between the two capabilities. Models trained on poorly grounded captions or with limited parameter updates fail to generalize, while high-quality long-caption data with strong visual grounding promotes both capabilities simultaneously. We further show that architectural choices aimed at preserving general alignment, such as frozen positional embeddings, can inadvertently limit compositional learning. Our analysis provides actionable guidelines for data selection and model design to improve VLM generalization.
\end{abstract}

\section{Introduction}

Understanding real-world images goes beyond the recognition of objects; it requires reasoning about their attributes and relationships within a scene. Captions that comprehensively describe such scenes are typically long, conveying not just more information but also greater compositional complexity. While vision-language models (VLMs) have made progress in understanding the relationship between images and text \citep{radford2021learning, jia2021scaling, chen2024spatialvlm}, their ability to interpret long, dense captions remains limited \cite{yamada2022lemons, kamath2023s, thrush2022winoground, garg2024imageinwords}. %

\begin{figure}[t]
  \centering  \includegraphics[width=\columnwidth]{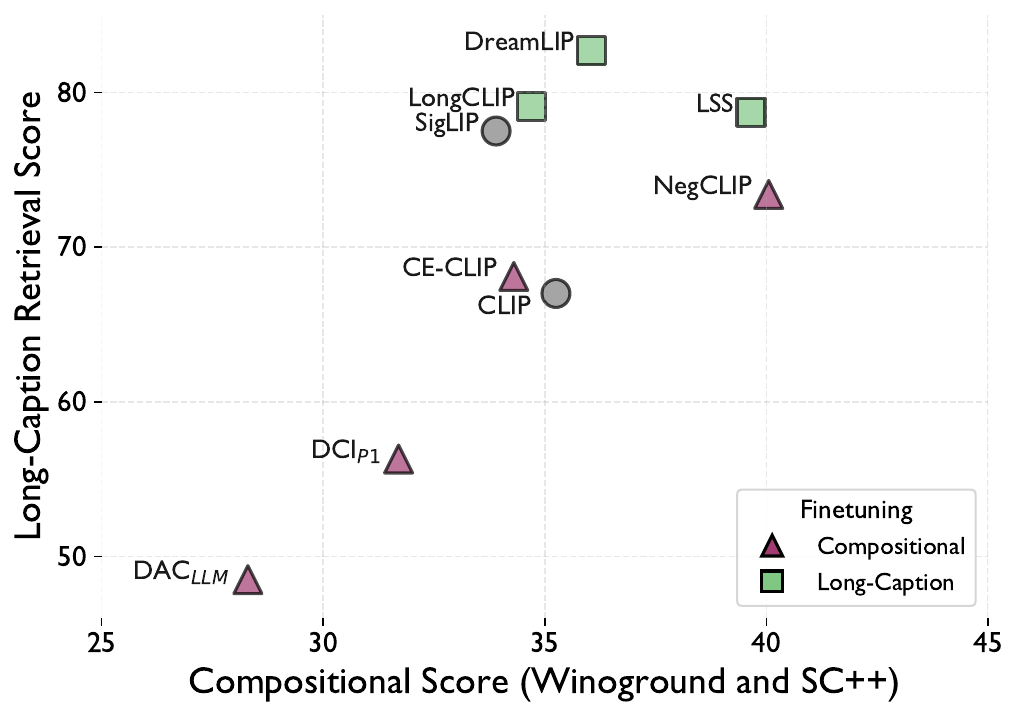}
  \caption{\textbf{Relationship between compositional reasoning and long-caption retrieval performance.} Models show a positive correlation between the two capabilities, but this relationship is sensitive to training setups, where architectural or optimization choices can limit compositional generalization.}
  \label{fig:correlation}
\end{figure}

Despite growing interest in both compositionality and long-caption understanding, current multimodal benchmarks typically treat these capabilities in isolation.
Compositionality has long been a focus of neural networks research \cite{lake2018generalization,bahdanau2019closure,hupkes2020compositionality}, but it has received limited attention in the vision-language literature~\cite{nikolaus2019compositional,suris2020learning}. More recently, several benchmarks have emerged to evaluate this ability in VLMs \cite{yuksekgonul2022and, thrush2022winoground, zhao2022explainable, hsieh2023sugarcrepe, dumpala2024sugarcrepe++}. However, these datasets often rely on short, generic captions and primarily test symbolic or relational binding, overlooking the linguistic and visual complexity present in real-world scenes. Meanwhile, benchmarks targeting long-caption understanding often use longer, dense captions \cite{cho2022fine, zhang2024long, urbanek2024picture, onoe2024docci, garg2024imageinwords}, but are rarely analyzed through the lens of compositionality \cite{urbanek2024picture}. As a result, the interplay between compositionality and long-caption understanding remains underexplored. %

Although compositionality is commonly viewed as both a prerequisite for and a consequence of long-caption understanding, models trained for long captions do not consistently exhibit strong compositional reasoning, and compositional models do not always generalize to long captions. We therefore ask two questions: (1) Does compositional training improve long-caption understanding? and (2) Does training on long, dense captions foster compositional generalization? To address these questions, we conduct a comparative analysis of models trained with distinct objectives, some targeting compositionality, others optimized for processing long captions. We evaluate them on a suite of benchmarks spanning compositional reasoning, long-caption retrieval (as a proxy for understanding dense descriptions), and general vision-language alignment, allowing us to disentangle the contribution of each skill and to assess whether gains transfer across tasks or remain narrowly scoped.

Our findings reveal a bidirectional relationship between compositionality and long-caption understanding, shaped by several important factors. Gains are sensitive to data quality: models trained on poorly structured or weakly grounded captions fail to generalize, while datasets with dense, diverse captions and broad vocabulary coverage lead to stronger performance across both tasks. We further identify trade-offs: Both compositional and long-caption training can degrade performance on general vision-language benchmarks, likely due to distributional shift. Likewise, training decisions that preserve general alignment, such as limited parameter updates or aggressive regularization, tend to underperform even when trained on high-quality data. 
We observe that transfer between compositional reasoning and long-caption understanding is highly sensitive to training design. In several cases, targeted improvements fail to transfer, and models exhibit structured trade-offs rather than uniform gains. However, even with these challenges, we show that models trained on well-designed long-caption data can achieve strong results in both long-caption retrieval and compositional reasoning.

\section{Background}
\label{sec:background}
Recent work has proposed architectural and objective modifications to improve VLM performance on long captions and on compositional tasks. CLIP~\cite{radford2021learning} offers strong general-purpose representations, but its short effective context window and weak compositional abilities expose clear limitations for complex captions~\cite{zhang2024long,kamath2023s,yamada2022lemons}. Recent efforts have introduced new training objectives \cite{yuksekgonul2022and, patel2024tripletclip} and improved data quality \cite{doveh2023dense, abbas2023semdedup} to boost performance.

\subsection{Compositional Reasoning}
\label{sec:background-compositionality}

\paragraph{Datasets}%
\citet{yuksekgonul2022and} showed that many standard VLM benchmarks can be solved by detecting objects alone, without modeling their relationships. They thus proposed ARO, a benchmark for relational and attribute understanding, built on Visual Genome~\citep{krishna2017visual}, using GQA~\citep{hudson2019gqa} annotations. Complementary work explored compositional challenges through controlled linguistic or visual perturbations. VL-Checklist \citep{zhao2022explainable}, and CREPE \citep{ma2023crepe} probe sensitivity to fine-grained structure such as object-attribute binding and spatial relationships, while Winoground~
\citep{thrush2022winoground} evaluates sensitivity to word order using contrastive image-caption pairs with identical lexical content but different semantic interpretations. However, it is known to entangle compositional reasoning with world knowledge \citep{diwan2022winoground}.

A key limitation of earlier benchmarks is their reliance on implausible or unnatural negatives, which can allow models to exploit superficial cues \citep{hsieh2023sugarcrepe}. To address this, the authors introduced SugarCREPE, with more plausible negatives generated by large language models, and SugarCREPE++ \citep{dumpala2024sugarcrepe++} further extends this benchmark with paraphrased positives to better assess compositional generalization.

\paragraph{Models}
Models targeting compositionality typically modify the training objective to encourage sensitivity to relational structure. NegCLIP~\citep{yuksekgonul2022and} and CE-CLIP~\citep{zhang2024contrasting} introduce hard or structured negative captions during contrastive training. DAC~\citep{doveh2023dense} and DCI~\citep{urbanek2024picture} focus on improving caption quality through denser or more localized descriptions. Additional implementation details are provided in Appendix~\ref{app:background-models}.

\subsection{Long-Caption Understanding}
\label{sec:background-long}
Standard CLIP-style models are trained on short, web-crawled captions and have a limited context window of 77 tokens, often effectively attending to only the first 20–30 tokens \citep{zhang2024long}. This motivates the use of long, dense captions both as a tool for training and for evaluating generalization under realistic linguistic complexity.

\paragraph{Datasets}
Several datasets provide long, dense captions that emphasize fine-grained descriptions, spatial relations, and contrastive distinctions. ImageInWords \citep{garg2024imageinwords} and DOCCI \citep{onoe2024docci} consist of human-written captions targeting detailed scene understanding, with DOCCI explicitly designed to distinguish between visually similar images. Urban1k \citep{zhang2024long} and DCI/sDCI \citep{urbanek2024picture}, derived from Visual Genome \citep{krishna2017visual}, use synthetically generated or summarized captions to capture complex object relations while remaining compatible with CLIP-style models.

Larger-scale efforts include Localized Narratives \citep{pont2020connecting}, which collects grounded human descriptions through synchronized narration and pointing, and synthetic approaches such as ShareGPT4V \citep{chen2024sharegpt4v} and LotLIP \citep{wu2024lotlip}, which generate long captions at scale using large language models applied to standard CLIP pretraining corpora. These datasets vary widely in scale, grounding, and caption structure, reflecting different trade-offs between annotation quality and dataset size.

\paragraph{Models}
Long-caption models have primarily focused on extending the effective textual context available during training. LongCLIP~\citep{zhang2024long} modifies the architecture to process longer input sequences, while DreamLIP~\citep{zheng2024dreamlip} emphasizes training on long, high-quality captions at scale without architectural changes.  Detailed model descriptions in Appendix~\ref{app:background-models}.

\begin{table*}[t]
\centering
\resizebox{\textwidth}{!}{%
\begin{tabular}{@{} l
    *{7}{S[table-format=2.1]} %
    c %
    *{9}{S[table-format=2.1]} %
    @{}}
\toprule
& \multicolumn{7}{c}{\textbf{Compositional Reasoning}}
&
& \multicolumn{9}{c}{\textbf{Long-Caption Retrieval}} \\
\cmidrule(lr){2-8} \cmidrule(lr){10-18}
& & \multicolumn{6}{c}{\textbf{SugarCrepe++}} &
& \multicolumn{2}{c}{\textbf{Urban1k}}
& \multicolumn{2}{c}{\textbf{sDCI}}
& \multicolumn{2}{c}{\textbf{DOCCI}}
& \multicolumn{2}{c}{\textbf{IiW}} & \\ %
\cmidrule(lr){3-8} \cmidrule(lr){10-11} \cmidrule(lr){12-13} \cmidrule(lr){14-15} \cmidrule(lr){16-17}
\textbf{Model} & {WG} & {SA} & {RR} & {RO} & {RA} & {SO} & {Avg.} %
& \multicolumn{1}{c}{} %
& {I2T} & {T2I} & {I2T} & {T2I} & {I2T} & {T2I} & {I2T} & {T2I} & {Avg.} \\ %
\midrule
CLIP & 17.2 & 39.1 & 47.4 & 85.3 & 62.4 & 32.5 & 53.3 & \multicolumn{1}{c}{} & 59.9 & 49.6 & 83.0 & 69.5 & 50.3 & 52.4 & 86.5 & 85.0 & 67.0 \\
SigLIP & \textbf{18.6} & 51.1 & 49.8 & 85.2 & 69.9 & 31.4 & 57.5 & & 62.9 & 62.3 & 88.0 & 78.0 & \textbf{70.3} & \textbf{70.8} & 93.0 & 94.5 & 77.5 \\
\cdashline{1-8}[0.5pt/1pt] \cdashline{10-18}[0.5pt/1pt] 
DAC$_{LLM}$  & 12.6 & 29.5 & 48.6 & 70.1 & 50.0 & 19.6 & 44.0 & \multirow{4}{*}{\boldmath{$\rightarrow$}} & 11.4 & 23.9 & 65.5 & 68.2 & 36.6 & 39.7 & 71.3 & 71.8 & 48.5 \\
DCI$_{P1}$   & 12.1 & 41.8 & 39.5 & 80.7 & 56.6 & 38.0 & 51.3 &  & 29.7 & 43.0 & 71.4 & 70.8 & 42.9 & 46.3 & 71.3 & 75.3 & 56.3 \\
CE-CLIP & 12.3 & 41.6 & 52.6 & 85.7 & 68.3 & 33.3 & 56.3 & & 53.5 & 65.0 & 82.0 & 74.8 & 43.1 & 55.4 & 73.3 & 85.3 & 68.1 \\
NegCLIP      & 16.4 & \textbf{57.5} & 52.0 & \textbf{92.1} & 73.0 & \textbf{43.9} & \textbf{63.7} & \multicolumn{1}{c}{} & 64.6 & 62.7 & 91.3 & 76.4 & 53.9 & 62.2 & 85.0 & 91.0 & 73.4 \\
\cdashline{1-8}[0.5pt/1pt] \cdashline{10-18}[0.5pt/1pt] %
LongCLIP & 14.7 & 40.8 & 48.4 & 89.1 & 65.6 & 29.6 & 54.7 & \multirow{3}{*}{\boldmath{$\leftarrow$}} & 77.9 & 77.7 & 86.4 & 74.6 & 61.4 & 69.3 & 90.8 & \textbf{95.0} & 79.1 \\
DreamLIP & 18.0 & 53.0 & 45.1 & 81.2 & 58.2 & 32.9 & 54.1 & & \textbf{79.8} & \textbf{79.6} & \textbf{94.7} & \textbf{82.0} & 69.9 & 69.5 & 93.0 & 93.0 & \textbf{82.7} \\
LSS & 17.5 & 52.2 & \textbf{53.4} & 91.3 & \textbf{74.9} & 36.5 & 61.8 &  & 75.4 & 74.1 & 91.7 & 75.1 & 64.5 & 63.0 & \textbf{94.0} & 92.0 & 78.7 \\
\bottomrule
\end{tabular}%
}
\caption{
\textbf{Compositional (left) and Long-Caption Retrieval (right) Performance Across Models.}
Compositional reasoning is evaluated on Winoground (WG, full results in Appendix \ref{app:winoground-groups}) and SugarCrepe++ with subcategories: SA (Swap Attribute), RR (Replace Relation), RO (Replace Object), RA (Replace Attribute), and SO (Swap Object). Long-caption retrieval is assessed on Urban1K, sDCI, DOCCI, and IiW using image-to-text (I2T) and text-to-image (T2I) metrics. \textbf{NegCLIP} leads on compositional tasks, while \textbf{LongCLIP-B} excels at long-caption retrieval. Arrows indicate cross-capability generalization: $\rightarrow$ shows gains in long-caption understanding from compositional training; $\leftarrow$ shows compositionality gains from long-caption training. %
}
\label{tab:merged-compositional-long}
\end{table*}

\section{Interplay of Compositionality and Long-Caption Understanding}
We investigate the relationship between compositional reasoning and long-caption understanding in vision-language models (VLMs). Specifically, we ask: (Q1) Does training for compositionality improve a model’s ability to interpret long, dense captions? and (Q2) Does training on long, structured captions promote compositional generalization? To answer these questions, we evaluate a set of off-the-shelf and custom-trained models using a unified benchmark suite, with all models tested in a zero-shot setting.

\subsection{Experimental Setup}
\label{sec:exper_setup}

\paragraph{Compositionality Benchmarks} We use Winoground and SugarCREPE++ (SC++) to evaluate compositionality. Although Winoground is an especially challenging benchmark, we include it as an upper bound for complex scene understanding, and, following \citet{diwan2022winoground}, we report their proposed grouped scores in Appendix~\ref{app:winoground-groups} to provide for fine-grained analysis. SC++, which encompasses and extends previous benchmarks, provides the most  robust evaluation of compositional generalization. We also discuss the ARO benchmark in \S~\ref{sec:dci-dac-fail-generalize}, but exclude VL-CheckList because some of its images are no longer available, preventing full reproducibility\footnote{This limitation has also been noted by~\citet{zhang2024contrasting}, who recommend avoiding VL-CheckList for new evaluations on their \href{https://github.com/lezhang7/Enhance-FineGrained}{project website}.}.

\paragraph{Long-caption Retrieval Benchmarks } To assess multimodal alignment under long, dense captions, we use zero-shot image-to-text and text-to-image retrieval as a proxy for understanding. We select datasets that contain detailed object-attribute bindings, spatial layouts, and nuanced descriptions beyond the scope of standard generic captions: Urban1K~\citep{zhang2024long}, sDCI~\citep{urbanek2024picture}, DOCCI~\citep{onoe2024docci}, and ImageInWords~\citep{garg2024imageinwords}. Dataset statistics, presented in Table~\ref{tab:dataset-stats}, highlight both the greater length of these captions and the increased complexity of their textual descriptions.

\paragraph{Main Models} To study the interplay between compositional training and long-caption understanding, we evaluate a set of
open-source VLMs. The compositional models we consider are NegCLIP, DAC$_{\text{LLM}}$, DCI$_{\text{P1}}$, and CE-CLIP, all trained with data and objectives designed to enhance compositional generalization (see Appendix~\ref{app:background-models}.) For long-caption understanding, we include LongCLIP-B and DreamLIP. As a baseline, we use the base CLIP model (ViT-B/32), which serves as the initialization for all other models in our study. We focus on CLIP-based models to enable controlled comparisons and isolate the effects of training data and objectives without architectural confounds.

\paragraph{Control Model} Most models in our study fine-tune CLIP (ViT-B/32) using diverse datasets and training objectives. However, LongCLIP modifies CLIP’s architecture to extend the input context, and DreamLIP uses a larger backbone (see \S\ref{sec:background-long}). These modifications may affect the interplay between long-caption training and compositionality in ways that are difficult to isolate. To control for these factors, we train the Long Story Short model (LSS), which finetunes CLIP (ViT-B/32) on the same ShareGPT4V data as LongCLIP, using standard contrastive loss and the original 77-token context window. This allows us to isolate the effect of long-caption data itself, independent of architectural changes, model size, or pretraining vs.\ finetuning. Importantly, LSS operates under the same architectural constraints as NegCLIP, DAC$_{\text{LLM}}$, and DCI$_{\text{P1}}$, enabling a direct and fair comparison between long-caption understanding and compositional training. We do not train LSS as a general-purpose model, but as a controlled intervention. Full details are in Appendix~\ref{app:training-parameters}.

\subsection{Does Compositional Training Improve Long-Caption Understanding?}
This question tests whether compositional properties evaluated by compositionality benchmarks are necessary for understanding long, dense captions with rich relational structure.

Table~\ref{tab:merged-compositional-long} reports compositional performance alongside downstream performance on long-caption retrieval for compositional models. We observe substantial variation among models on compositional benchmarks. DAC and DCI exhibit poor performance on both SC++ and WG, falling short even of the base CLIP model. Only NegCLIP consistently achieves a sizeable improvement over CLIP across all SC++ subcategories while maintaining a reasonable score on Winoground, indicating that contrastive training augmented with hard negatives can effectively induce compositionality.

Turning to long-caption retrieval, we observe a closely aligned pattern. NegCLIP, despite being trained only on short COCO captions, substantially outperforms CLIP across all long-caption benchmarks and closes much of the gap to LSS and LongCLIP. CE-CLIP exhibits intermediate transfer, while DAC and DCI fail to generalize, consistently underperforming even CLIP. Overall, long-caption retrieval performance mirrors compositional performance, with NegCLIP achieving the strongest results among the compositional models.

This parallel ordering across SC++ and long-caption retrieval provides strong evidence that \textbf{compositional reasoning generalizes to and supports long-caption understanding}. We further analyze the failure modes of DAC and DCI in Section~\ref{sec:dci-dac-fail-generalize}. Having established the relationship between compositionality and long caption understanding in one direction, we turn to our second research question.

\subsection{Do Long Captions Promote Compositionality?}
\label{sec:long-to-comp}

The basis for this question is two-fold: (1) long, detailed captions provide a rich signal for learning fine-grained vision-language alignment, and (2) their richer compositional structure can foster generalizable compositional abilities.

Table~\ref{tab:merged-compositional-long} presents the results. We first consider long-caption retrieval performance for LongCLIP, DreamLIP, and our control model, LSS. While performance varies between the retrieval datasets, all three models perform strongly. LSS, despite retaining CLIP's original 77-token context window, outperforms LongCLIP on several subtasks and falls short by only 0.4 points on average, suggesting limited benefit from LongCLIP’s extended context. DreamLIP achieves the strongest overall retrieval performance, consistent with its larger backbone and full pretraining on long captions.%

Next, we consider the performance of these models on compositionality benchmarks. LSS substantially improves over CLIP on SC++ and Winoground, nearly matching NegCLIP’s average performance and surpassing it in some SC++ subcategories. LSS also achieves a high Winoground score, providing strong evidence that \textbf{training on long, grounded captions promotes compositional generalization}. DreamLIP performs well on WG and shows consistent, though smaller, gains over CLIP on SC++, while its larger model size and full pretraining introduce confounding factors.

In contrast, LongCLIP shows little improvement over CLIP on SC++ or Winoground, despite being trained on the same ShareGPT4V data as LSS. This divergence suggests that architectural constraints, specifically freezing early positional embeddings, can limit gains in compositional reasoning. We further analyze this effect in \S~\ref{sec:longclip-analysis}. Figure~\ref{fig:correlation} summarizes these results.

\section{Limits of the Bidirectional Relationship}

This section presents further analysis of the results discussed above, focusing on cases where the bidirectional relationship between compositionality and long-caption understanding does not hold uniformly. While these capabilities can reinforce each other, we observe substantial variation across models driven by data quality, training dynamics, and architectural constraints. We analyze these failure modes and examine how gains in compositional and long-caption understanding interact with other core vision-language capabilities.

\begin{table}[t]
\centering
\resizebox{\columnwidth}{!}{%
\begin{tabular}{@{} l
*{5}{S[table-format=2.1]} 
@{}}
\toprule
& \multicolumn{4}{c}{\textbf{ARO Benchmark}} \\
\cmidrule(lr){2-5} %
\textbf{Model} & {VG-R} & {VG-A} & {COCO} & {Flickr} & \textbf{SC++}\\
\midrule
CLIP & 59.8 & 63.0 & 47.3 & 58.5 & 53.3\\
SigLIP & 34.8 & 55.9 & 32.7 & 40.7 & 57.5 \\
\cdashline{1-6}[0.5pt/1pt]
DAC$_{LLM}$ & 81.3 & 73.9 & \textbf{94.5} & \textbf{95.7} & 44.0 \\
DCI$_{P1}$ & 72.6 & 67.6 & 88.6 & 91.3 & 51.3 \\
CE-CLIP & 81.2 & 75.6 & 71.9 & 75.6 & 56.3 \\
NegCLIP& \textbf{81.8} & 72.1 & 82.5 & 86.7 & \textbf{63.7}\\
\cdashline{1-6}[0.5pt/1pt]
LongCLIP & 59.7 & 63.4 & 56.9 & 69.0 & 54.7 \\
DreamLIP & 51.2 & \textbf{79.0} & 52.0 & 49.8 & 54.1\\
LSS & 62.0 & 65.7 & 37.7 & 46.0 & 61.8\\
\bottomrule
\end{tabular}%
}
\caption{\textbf{Compositional Reasoning Performance on ARO and SugarCrepe++.}
Compositional reasoning performance of various models on the ARO benchmark and the average SugarCrepe++ (SC++) score. Notably, models like DAC$_{LLM}$ and DCI$_{P1}$ almost saturate the ARO benchmark, yet their performance on SC++ often shows a poor correlation, indicating ARO's limitations in evaluating modern compositional abilities.
}
\vspace{-1em}
\label{tab:results-aro}
\end{table}

\subsection{The Case of DAC and DCI}
\label{sec:dci-dac-fail-generalize}
The DAC$_\text{LLM}$ and DCI$_\text{P1}$ models were trained with compositional objectives and hard negatives on long-caption data (see Appendix ~\ref{app:background-models}.), but they consistently underperform on the SC++ and Winoground benchmarks and on long-caption retrieval tasks (see Table~\ref{tab:merged-compositional-long}.) This failure contrasts sharply with the success of NegCLIP and LSS, which each use one component from this training recipe and exhibit strong generalization.  

The original evaluations of DAC and DCI relied heavily on ARO, a legacy benchmark consisting of constrained, rule-based captions. As shown in Table~\ref{tab:results-aro}, both models perform well on ARO, nearly saturating the benchmark. However, our analysis reveals a negative correlation between ARO and SC++ performance (Spearman $r=-0.37$), exposing a disconnect between ARO and more challenging contemporary evaluations. SC++ includes more natural distractors and paraphrased positives, requiring generalizable compositional reasoning. Strong ARO performance appears to reflect optimization for a narrow metric rather than true generalization. Still, this mismatch alone does not fully explain why DAC and DCI perform so poorly on the tasks they were trained to address.

\begin{figure*}[t]
  \includegraphics[width=\textwidth]{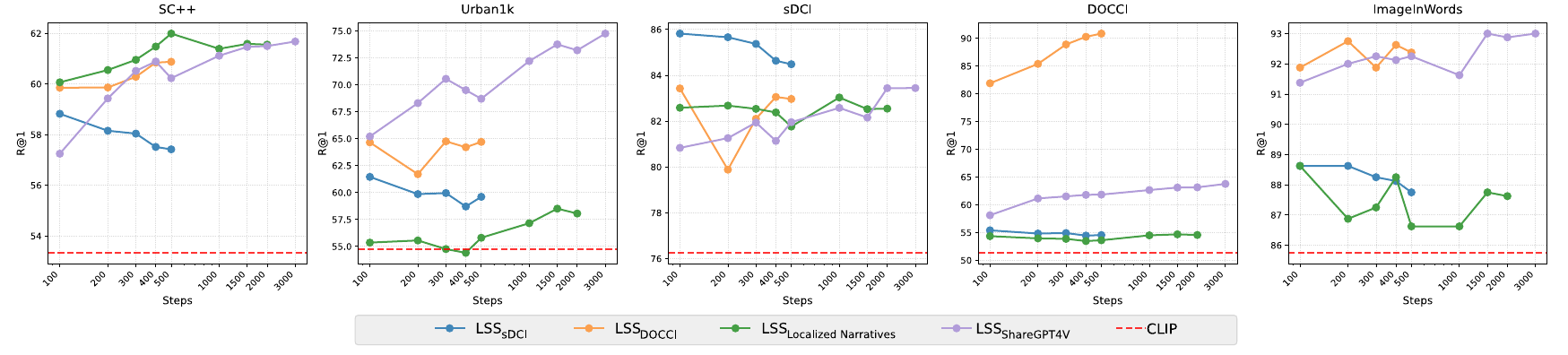}
  \caption{\textbf{Training Dynamics Across Long-Caption Datasets}. Evolution of performance for models trained on four long-caption datasets. Results are reported for both long-caption retrieval (Urban1K, DOCCI, sDCI, IiW) and compositional benchmark, SC++. Models trained on ShareGPT4V and DOCCI show consistently stronger generalization, highlighting the role of caption quality and grounding beyond dataset scale.}
  \label{fig:training-dynamics}
\end{figure*}

We posit that the failure to generalize stems from two core issues. First, the training captions used by DAC and DCI may be of low quality. DAC and DCI rely heavily on synthetic captions, either blind LLM-generated expansions or disjoint region-level descriptions, that may often lack fluency, cohesion, or visual grounding. These deficiencies limit the ability of a model to learn grounded, compositional alignments—even in the presence of hard negatives and contrastive loss. Second, both DAC and DCI apply LoRA-based updates to only a limited subset of model parameters. This lightweight adaptation restricts the models’ capacity to internalize compositional structure beyond surface-level patterns. In contrast, NegCLIP and LSS were fully finetuned, allowing deeper representational shifts. 

Together, these findings highlight the importance of not just what models are trained on, but also how they are trained. Compositional objectives and long captions alone are not sufficient; sufficient parameter adaptation and data quality are critical to realizing the intended generalization. Next, we examine the role of the training data further.

\subsection{Are All Long-Caption Datasets the Same?}
\label{sec:all-datasets}

To study how different types of long-caption data affect compositional reasoning and long-caption retrieval, we train additional LSS models on the sDCI Train, DOCCI Train, Localized Narratives datasets, using the same experimental setup from \S\ref{sec:exper_setup}). These datasets vary in scale and linguistic complexity, range from 7K to over 1M examples in size, and span both synthetic and human-written captions. Table~\ref{tab:train-dataset-stats} shows detailed dataset statistics. This controlled intervention allows us to systematically examine how properties like grounding, compositional complexity, and size influence generalization.

We show the results in Figure~\ref{fig:training-dynamics} (see Appendix~\ref{app:LLS-results}). all LSS models outperform the CLIP baseline on long-caption retrieval and exhibit some degree of compositional transfer. However, performance varies notably: models trained on ShareGPT4V and DOCCI consistently outperform those trained on sDCI and Localized Narratives, both in the course of training and at convergence. This indicates that generalization depends not just on scale, but also on the quality and structure of the training data. We now discuss each training dataset in turn. 

\noindent{\textbf{ShareGPT4V.}}
LSS$_\text{ShareGPT4V}$ achieves the strongest results overall, excelling in both long-caption retrieval and compositional benchmarks like SC++. This success stems from a combination of favorable dataset properties (Table~\ref{tab:dataset-stats}) rather than any single factor. At 1.2M examples, ShareGPT4V is the largest scale dataset, but more critically, it achieves near-complete vocabulary coverage (87.72\% of CLIP's tokenizer vocabulary), far exceeding the $\sim$25\% coverage of the other datasets. The average caption lengths are far-beyond the 77-token limit of CLIP, ensuring the model trains the full range of positional embeddings. Interestingly, ShareGPT4V exhibits only moderate syntactic complexity (Yngve depth: 45.70), which is significantly lower than the other datasets. This suggests that the combination of scale, comprehensive vocabulary coverage, and appropriate caption length compensates for moderate structural complexity, enabling robust generalization on both compositional and long-caption benchmarks.

\vspace{3mm} \noindent{\textbf{DOCCI.}} Despite the small size of DOCCI (14.6K captions), LSS$_\text{DOCCI}$ achieves strong performance comparable to LSS$_\text{ShareGPT4V}$ across long-caption retrieval tasks. As shown in Table~\ref{tab:dataset-stats}, its strength lies in being human-written, visually grounded, and contrastive by design, explicitly crafted to distinguish between similar images. With high syntactic complexity (Yngve: 74.55) and long captions (122 tokens), DOCCI demonstrates that careful curation and purposeful annotation strategies can rival large-scale data collection, achieving comparable vocabulary coverage (26.96\%) despite its size.

\vspace{3mm}
\noindent{\textbf{sDCI.}}
LSS$_\text{sDCI}$ achieves strong performance on the sDCI test set, but performs less consistently on other benchmarks. Although the sDCI train set shows the highest syntactic complexity (Yngve: 94.07), this may stem from LLM-generated caption summarized from multiple captions exceeding 1000 words, which may artificially inflate syntactic complexity without good semantic coherence or visual grounding. Its small image set (7.6K unique images) may also lead to overfitting, as indicated by the declining performance on SC++ and Urban1K with more training steps. This highlights that complexity alone is not enough without grounding and sufficient visual diversity.

\vspace{3mm}
\noindent{\textbf{Localized Narratives.}}
With 489K captions, LN is the only large-scale human-annotated dataset available. LSS$_\text{LN}$ shows the fastest early gains on SC++ among the long-caption datasets (Figure~\ref{fig:training-dynamics}), likely thanks to a moderate-high Yngve complexity (61.70). However, the model lags behind on all the long-caption retrieval benchmarks. This can be attributed to caption length: LN captions average only 30 words (Table~\ref{tab:dataset-stats}), well below CLIP’s 77-token limit and far shorter than the other DOCCI, sDCI and ShareGPT4V. Moreover, LN has the lowest vocabulary coverage (24.34\%). LN provides enough signal to improve compositionality, but the captions are too short on average to support strong long-caption understanding.

Our analysis shows that dataset effectiveness depends on the interaction of multiple properties---scale, vocabulary coverage, caption length, syntactic complexity, and annotation quality---rather than any single factor. High-performing datasets like ShareGPT4V and DOCCI succeed through complementary strengths: large scale and broad coverage in the former, careful grounding and contrastive structure in the latter. In contrast, sDCI and Localized Narratives highlight key pitfalls: overly complex or synthetic captions without grounding, and human-annotated but short captions with limited vocabulary. Ultimately, strong generalization emerges not from any one property, but from a balanced design that aligns linguistic richness with visual grounding and sufficient sequence length.

\begin{table}[t]
\centering
\resizebox{\columnwidth}{!}{%
\begin{tabular}{@{} lccc
S[table-format=2.2]
@{}}
\toprule
\textbf{Dataset} & \textbf{Images} & \textbf{Captions} & \textbf{Avg. Length} & {\textbf{Vocab.}} \\
   &  &  &  & \textbf{Covered (\%)} \\
\midrule
sDCI$_{\text{Train}}$   & $7.6{\times}10^3$   & $8.3{\times}10^4$   & 40 $\pm$ 12  & 29.29 \\
DOCCI$_{\text{Train}}$ & $1.5{\times}10^4$   & $1.5{\times}10^4$   & 122 $\pm$ 45 & 26.96 \\
LN                     & $4.9{\times}10^5$   & $4.9{\times}10^5$   & 30 $\pm$ 17  & 24.34 \\
ShareGPT4V             & $1.2{\times}10^6$   & $1.2{\times}10^6$   & 144 $\pm$ 39 & 87.72 \\
\bottomrule
\end{tabular}
}
\caption{\textbf{Training dataset statistics.} Dataset size, average caption length, and vocabulary coverage with respect to CLIP’s tokenizer. ShareGPT4V stands out for its scale, long captions, and near-complete vocabulary coverage. Additional dataset statistics in Table~\ref{tab:dataset-stats} (including syntactic complexity, word ranges, source types, and evaluation datasets).}
\label{tab:train-dataset-stats}
\end{table}

\subsection{The Case of LongCLIP}
\label{sec:longclip-analysis}
Our main results in Table~\ref{tab:merged-compositional-long} revealed a surprising finding: despite having a very different context window size, LSS$_{ShareGPT4V}$ and LongCLIP, trained on the same ShareGPT4V dataset, show comparable performance on long-caption retrieval, and the former even outperforms the latter on compositionality tasks. At first sight, this is counterintuitive, since LongCLIP is expected to benefit from a major architectural advantage, which allows the model to process up to 248 tokens (compared to 77 for LSS.) However, the differences are not just in architecture but also in training procedure. LSS operates within CLIP’s original 77-token limit, updating all parameters during training. LongCLIP, by contrast, uses an extended context window but freezes the first 20 positional embeddings and applies reduced updates to positions 20 to 77. These modifications aim to preserve the strong generalization capabilities of the base CLIP model. Yet, they may also interfere with the learning of new skills in this part of the context window.

\begin{figure}[t]
  \includegraphics[width=\linewidth,height=150pt]{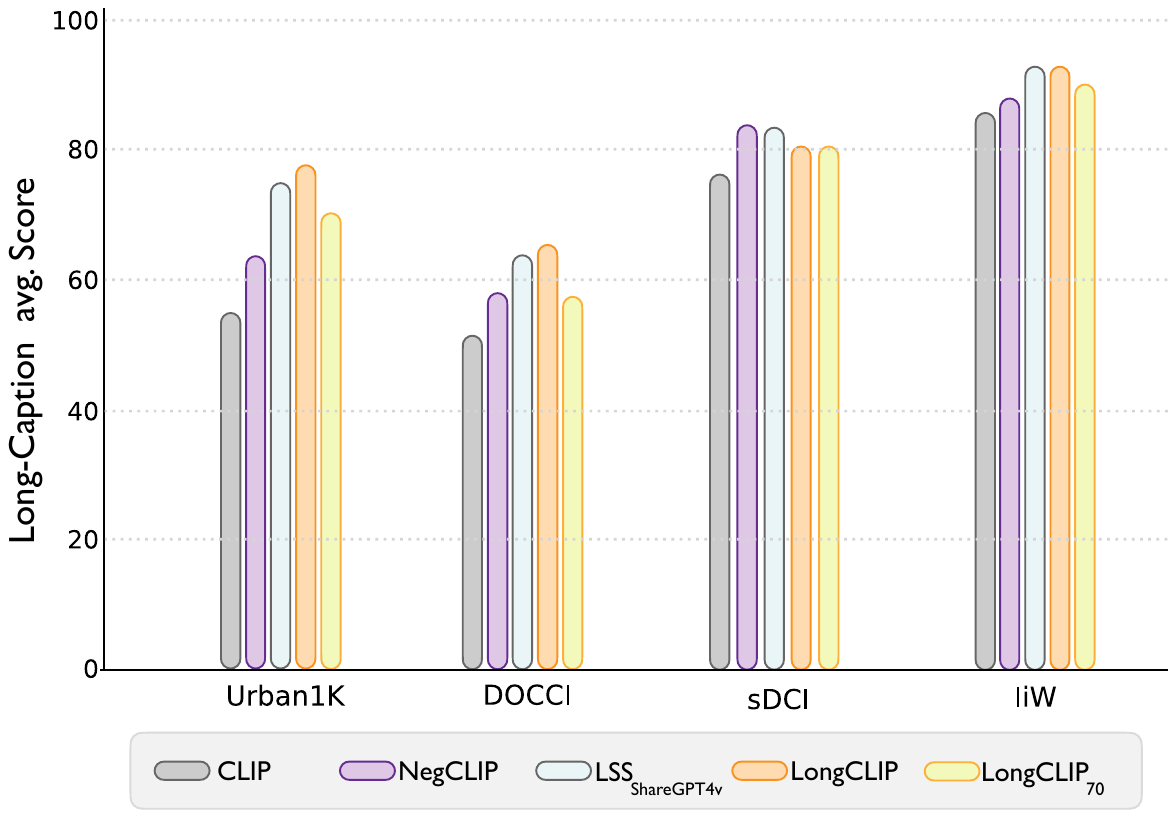}
  \caption{\textbf{Long-caption Retrieval Performance}. LSS achieves strong performance on all benchmarks, with LongCLIP outperforming by a small margin despite its three times larger input token processing capacity. When truncating LongCLIP to 70 words, LSS closes the gap and outperforms it across all benchmarks. These results demonstrate that full parameter adaptation enables more effective long-caption understanding.}
  \label{fig:bar-performance}
\end{figure}

To isolate this effect, and eliminate the confounding factor of longer input length, we evaluate a truncated version of LongCLIP: LongCLIP$_{70}$, which limits inputs to 70 words (approximately $\sim$77 tokens). As shown in Figure~\ref{fig:bar-performance}, performance on long-caption retrieval for LongCLIP$_{70}$ drops sharply, and LSS$_\text{ShareGPT4V}$ now dominates on both compositionality and long-caption understanding.
The main limitation of LongCLIP thus appears to stem from its training constraints. Freezing the first 20 positional embeddings on the assumption that they are already well-optimized for vision–language alignment, limits the ability to learn new relational patterns. As a result, it preserves a bag-of-words–like processing, particularly in the early input positions where most compositional benchmarks operate. In the next section, we explore this trade-off in more depth.

\subsection{Trade-offs for Better Compositionality and Long-Caption Understanding}
To better understand how compositionality and long-caption training affect general alignment, we evaluate all models on standard vision-language benchmarks.
We assess classification on CIFAR10, CIFAR100 and ImageNet using prompt templates from prior work~\citep{radford2021learning}, and
retrieval (measured in Recall@1) on the COCO and Flickr30k test sets. Results are shown in Table~\ref{tab:general-capabilities}.

LongCLIP achieves the strongest overall performance, particularly on classification and short-caption retrieval. This attests to the effectiveness of preserving CLIP's alignment via the freezing of positional embeddings. Since many COCO and Flickr30k captions exceed CLIP's functional 20-token window (Table~\ref{tab:dataset-stats}), LongCLIP gains a natural retrieval advantage.
NegCLIP achieves the highest recall on most short-caption retrieval tasks, demonstrating that compositional reasoning can enhance general vision-language understanding. However, this advantage may partly reflect in-distribution bias, as NegCLIP is trained on COCO, which appears in the evaluation set. Despite strong retrieval performance, NegCLIP shows degraded classification accuracy compared to base CLIP. We observe the same pattern in our LSS$_{\text{ShareGPT4V}}$ model: it consistently improves over base CLIP on all retrieval tasks but underperforms on classification, particularly on ImageNet. A likely explanation is the mismatch between training and evaluation distributions---these models are trained on longer, relational captions, but evaluated using fixed, template-based prompts (e.g., “a photo of a dog”), which favor alignment patterns seen during CLIP’s original training.

\begin{table}[t]
\centering
\resizebox{\columnwidth}{!}{%
\begin{tabular}{@{} l *{3}{S[table-format=2.1]} *{4}{S[table-format=2.1]} @{}}
\toprule 
& \multicolumn{3}{c}{\textbf{Zero-Shot Classification}} & \multicolumn{4}{c}{\textbf{Retrieval}} \\
\cmidrule(lr){2-4} \cmidrule(lr){5-8}
 & & & & \multicolumn{2}{c}{COCO} & \multicolumn{2}{c}{Flickr30k} \\
\cmidrule(lr){5-6} \cmidrule(lr){7-8}
 \textbf{Model} & \textbf{C10} & \textbf{C100} & \textbf{IN1K} & {I2T} & {T2I} & {I2T} & {T2I} \\
\midrule
CLIP       & 89.8 & 65.1 & 63.1 & 50.4 & 30.2 & 78.6 & 59.0 \\
SigLIP & 92.4 & \textbf{72.3} & \textbf{76.0} & \textbf{65.7} & \textbf{47.8} & \textbf{88.9} & 74.7\\
\cdashline{1-8}[0.5pt/1pt]
DAC$_{\text{LLM}}$ & 90.4 & 64.1 & 51.1 & 33.7 & 37.7 & 53.1 & 64.9 \\
DCI$_{\text{P1}}$  & 87.1 & 58.0 & 53.3 & 20.5 & 21.4 & 55.9 & 44.0 \\
CE-CLIP & 85.9 & 60.2 & 50.0 & 55.3 & 46.9 & 74.9 & 68.3 \\
NegCLIP        & 88.9 & 63.2 & 61.0 & 59.3 & 44.8 & 85.1 & 70.9 \\
\cdashline{1-8}[0.5pt/1pt]
LongCLIP     & 91.3 & 69.5 & 66.9 & 57.2 & 40.6 & 86.2 & 70.7 \\
DreamLIP & \textbf{92.7} & 67.0 & 55.7 & 57.6 & 47.7 & 84.7 & \textbf{75.0} \\
LSS & 88.9 & 65.8 & 60.8 & 57.2 & 38.9 & 83.0 & 68.4 \\
\bottomrule
\end{tabular}%
}
\caption{\textbf{Zero-shot Performance on General Vision-Language Tasks.} We report classification accuracy on CIFAR10 (C10), CIFAR100 (C100), and ImageNet-1K (IN1K), and Recall@1 for image-text retrieval on COCO and Flickr. LongCLIP performs strongly across all tasks, indicating that freezing initial positional embeddings preserves general alignment and even enhances it through the remaining positions. While NegCLIP and LSS underperform on classification, both achieve strong retrieval scores.}
\label{tab:general-capabilities}
\end{table}

These results point to a nuanced trade-off in VLM design. Constraints like frozen positional embeddings, like those used in LongCLIP, can preserve general performance, and even support improved performance when combined with architectural modifications or high-quality data. Compositional and long-caption understanding, as seen in LSS and NegCLIP, boosts retrieval but tends to degrade classification accuracy. This indicates that while compositionality and long-caption understanding benefit each other, both properties may come at the cost of general-purpose alignment, highlighting the importance of aligning training strategies with intended use cases.

\section{Conclusion}
We investigated the relationship between compositional reasoning and long-caption understanding in contrastive vision-language models. Our results reveal a bidirectional link: compositional training improves performance on long-caption retrieval, while training on long, complex captions fosters compositional generalization. However, we find that neither compositional objectives nor long captions are sufficient on their own. Robust generalization emerges only when combined with sufficient parameter adaptation and high-quality training data consisting of grounded, comprehensive captions with syntactic and lexical diversity. 

Compositional generalization, whether achieved via targetted objectives or training data, is key to developing models that generalize across caption complexity. As models are increasingly applied to open-ended, real-world tasks, supporting robust generalization across diverse and structured language should be a central goal---one that requires choosing the right benchmarks, model design and training strategies.

\section*{Limitations}
Our work highlights the benefits of long-caption training but focuses primarily on general performance, compositionality, and long-caption retrieval on contrastive visual-language models. We do not evaluate generative VLMs, which involve additional factors such as autoregressive decoding, alignment procedures, and cross-attention dynamics.
Complex caption phenomena such as temporal or causal reasoning are not addressed. We also rely on long-caption retrieval as a proxy for understanding; more targeted benchmarks are needed for deeper analysis, potentially involving generative evaluation or fine-grained VLM probing. Finally, we do not explore combinations of training losses or architectural modifications; nonetheless, our results indicate that substantial gains are possible even within the original CLIP input constraints. 
While we observe strong correlations between compositional training and long-caption understanding, our analysis does not fully isolate the underlying causal mechanisms. Future work should investigate the specific roles of vocabulary coverage, syntactic complexity, and visual grounding through targeted mechanistic interventions.

\section*{Acknowledgments}
This work was supported by research grant (VIL53122) from Villum Fonden. We acknowledge the EuroHPC Joint Undertaking for awarding this project access to the EuroHPC supercomputer LEONARDO, hosted by CINECA (Italy) and the LEONARDO consortium through an EuroHPC Development Access call (ID:EUHPC\_D12\_071). 

\bibliography{custom}

\appendix
\section{General-Purpose VLM Evaluation}
\label{general-vlm-evaluation}
Pretraining VLMs using image-text pairs and contrastive loss functions~\cite{oord2018representation} has proven highly effective for learning aligned multimodal representations \cite{radford2021learning,jia2021scaling,pham2023combined}, demonstrating strong generalization in zero-shot settings, where no task-specific finetuning is applied.
Pretrained models are commonly evaluated on zero-shot classification  datasets like CIFAR~\cite{krizhevsky2009learning}, as a benchmark for coarse-grained object recognition, and ImageNet-1k~\cite{russakovsky2015imagenet}, evaluating fine-grained classification across 1,000 categories. Oxford-IIIT Pets~\cite{parkhi2012cats}, Stanford Cars~\cite{krause20133d}, and Food101~\cite{bossard2014food} focus on fewer but more specific classes. Cross-modal retrieval benchmarks, such as MS-COCO~\cite{lin2014microsoft} and Flickr30K~\cite{plummer2015flickr30k}, contain short and general captions,  evaluating alignment between visual and textual modalities through image-to-text and text-to-image matching. However, these benchmarks primarily test coarse-level alignment or object recognition, offering limited insight into compositional understanding.

\section{Compositional and Long-Caption Models}
\label{app:background-models}
\paragraph{Compositional Models} 
Various data augmentations and objective modifications have been proposed to improve the compositionality of VLMs. NegCLIP~\cite{yuksekgonul2022and} extends the contrastive loss with hard negative captions, encouraging finer discriminative alignment. Hard negatives are mined through nearest-neighbour search and synthetically created through word order perturbations. Densely Aligned Captions \cite[DAC]{doveh2023dense} targeted both data quality and training objectives. Based on the intuition that more descriptive captions can enhance compositionality, they introduced two methods for rewriting CC3M captions into more descriptive and semantically rich versions: 1) using an LLM to expand short captions with plausible scene-level details, and 2) using SAM~\cite{kirillov2023segment} to segment objects and generate short captions for each region. 
Two models were trained, DAC$_{\text{LLM}}$ and DAC$_{\text{SAM}}$, using LoRA\cite{hu2022lora}, with DAC$_{\text{LLM}}$ performing better on ARO. %
\citet{urbanek2024picture} introduced a human-annotated dataset in which annotators were asked to describe automatically segmented regions of an image in detail. These region-level captions were then summarized to fit within the context window of CLIP, and used to fine-tune a vision-language model. They trained DCI$_{\text{P1}}$ with both contrastive and negative loss showing strong performance on ARO and VL-Checklist. CE-CLIP~\cite{zhang2024contrasting} expands the hard-negative approach for compositionality by generating multiple targeted negatives for each image, focusing on relationships, attributes, and actions. The model is trained with the standard contrastive loss plus two additional objectives: (1) an intra-modal contrastive loss comparing correct captions to hard negatives, and (2) a cross-modal contrastive loss enforcing higher similarity between image–correct pairs than image–negative pairs. SPARCL~\cite{li2025enhancing}, in contrast, introduces both positive and negative captions and images using a text-to-image generative pipeline.\footnote{Weights were not available at the time of writing.}

\paragraph{Long-Caption Models}

LongCLIP~\cite{zhang2024long} focuses on long-caption processing by extending the textual context window from 77 to 248 inputs by interpolating positional embeddings. To retain the strengths of the original CLIP, the first 20 positional embeddings are frozen, while the remaining base encoder up to the token 77 are interpolated with a heavier weighting toward the pretrained positions. The rest of the model is fully trained from a CLIP initialization.
DreamLIP~\cite{zheng2024dreamlip} explores the impact of training with long, high-quality captions. The authors recaption 30M images and train using a global multi-positive contrastive loss, pairing each image with multiple subcaptions, and a fine-grained loss aligning image subpatches with their corresponding subcaptions. Unlike other models discussed in this paper, DreamLIP uses a larger CLIP backbone (ViT-B/16 rather than ViT-B/32) and conducts a data-scaling study, training multiple model variants with increasing data sizes.

\section{Training Parameters}
\label{app:training-parameters}
In this section, we present the training parameters for our models. Models are named after the datasets on which they were trained.

We fine-tune CLIP models on each dataset, fixing the batch size to 1024 for all runs and adjusting the number of training steps accordingly. Training is performed using 4× A100 GPUs to accommodate the ViT-B/32 version of CLIP from HuggingFace.\footnote{Pretrained models will be released under an open-source license.} Visual and textual input processing follows the default parameters of the pretrained CLIP models, as specified in the HuggingFace model card. The longest training run required 8 hours. Detailed training parameters are shown in Table \ref{tab:parameters}.

\begin{table}[t]
\centering
\resizebox{\columnwidth}{!}{%
\begin{tabular}{lrrrrr}
\toprule
\textbf{Model} & \textbf{WarmUp} & \textbf{LR} & \textbf{Steps} & \textbf{Epochs} & \textbf{Checkpoints} \\
\midrule
sDCI & 5 & 5e-6 & 500 & 70 & [100, 200, 300, 400, 500] \\
DOCCI & 15 & 5e-6 & 500 & 35 & \textit{PREV} \\
LN & 50 & 3e-6 & 2000 & 4 & [\textit{PREV}, 1000, 1500, 2000] \\
ShareGPT4V & 150 & 3e-6 & 3000 & 2.5 & [\textit{PREV}, 3000]\\
\bottomrule
\end{tabular}
}
\caption{\textbf{Training Parameters.} \textit{PREV} includes all checkpoints from the model on the previous row, and \textit{LR} denotes the learning rate. We report the number of training steps and their approximate equivalent in epochs.}
\label{tab:parameters}
\end{table}

\begin{table}[b]
\centering
\resizebox{\columnwidth}{!}{%
\begin{tabular}{@{} l
  *{8}{S[table-format=2.1]}
  @{}}
\toprule
& \multicolumn{3}{c}{\textbf{Winoground}}
& \multicolumn{4}{c}{\textbf{Long Retrieval}} \\
\cmidrule(lr){2-4} \cmidrule(lr){5-8}
\textbf{Model} & {Group} & {Image} & {Text} & {Urban} & {sDCI} & {DOCCI} & {IiW} \\
\midrule
Siglip & 10.3 & 12.8 & 32.8 & 62.6 & 83.2 & 70.6 & 93.8 \\
DreamLIP & 10.8 & 15.0 & 28.3 & 79.7 & 88.3 & 69.7 & 93.0 \\
LSS (ViT-B/16) & 8.8 & 11.3 & 30.8 & 81.3 & 83.7 & 67.8 & 94.8 \\
\midrule
& \multicolumn{3}{c}{\textbf{Classification}}
& \multicolumn{2}{c}{\textbf{Short Retrieval}}
&
& 
& \\
\cmidrule(lr){2-4} \cmidrule(lr){5-6}
 & {CIFAR10} & {CIFAR100} & {ImageNet} & {COCO} & {Flickr30k} & {SC++} \\
\midrule
Siglip & 92.4 & 72.3 & 76.0 & 56.8 & 81.8 & 57.5 \\
DreamLIP & 92.7 & 67.0 & 55.7 & 52.7 & 79.9 & 54.1\\
LSS (ViT-B/16) & 91.4 & 68.0 & 65.1 & 51.8 & 81.2 & 60.4 \\
\bottomrule
\end{tabular}
}
\caption{
\textbf{Generalization Across Architectures.} DreamLIP (pretrained on long captions) and LSS/16 (fine-tuned on long captions) achieve comparable performance across benchmarks. SigLIP, despite lacking explicit long-caption training, matches these models on long-caption retrieval and performs strongly on compositional benchmarks. These results reinforce that the observed relationship between long-caption understanding and compositionality generalizes across architectures and training paradigms.
}
\label{tab:architecture-variation}
\end{table}
\section{Dataset Statistics}

Table~\ref{tab:dataset-stats} presents an overview of the datasets used in this study, in terms of the caption lengths, vocabulary coverage, and textual complexity, as estimated by Yngve metric.

\section{Pretraining vs. Fine-tuning}
\label{app:pretrain-vs-finetune}
DreamLIP provides an opportunity to compare the effects of full pretraining with long captions against finetuning on long-caption data. Although the setups are not perfectly aligned, we fine-tune a larger CLIP model (ViT-B/16) using the same training recipe as LSS$_{\text{ShareGPT4V}}$. The results are summarized in Table~\ref{tab:architecture-variation}.

We observe comparable performance between DreamLIP and our finetuned model on retrieval tasks. However, they differ notably on other benchmarks. DreamLIP achieves the strongest Winoground score overall, whereas finetuning on a larger backbone does not significantly boost Winoground performance. In contrast, our fine-tuned LSS/16 model outperforms DreamLIP on SC++. The bigger differences emerge in image classification. Our fine-tuned model achieves the highest classification accuracy among all evaluated models, whereas DreamLIP underperforms even CLIP-B/32 despite its larger backbone. This may reflect the fact that pretraining exclusively on long captions introduces a distribution shift that harms zero-shot classification performance. For short retrieval, both models perform similarly.

\begin{table*}[b]
\centering
\resizebox{\textwidth}{!}{%
\begin{tabular}{@{} l
  *{27}{S[table-format=2.1]} %
  @{}}
\toprule
& \multicolumn{3}{c}{\textbf{Overall}}
& \multicolumn{3}{c}{\textbf{Non Compositional}}
& \multicolumn{3}{c}{\textbf{Unusual Image}}
& \multicolumn{3}{c}{\textbf{Visually Difficult}}
& \multicolumn{3}{c}{\textbf{Unusual Text}}
& \multicolumn{3}{c}{\textbf{Ambiguously Correct}}
& \multicolumn{3}{c}{\textbf{Complex Reasoning}}
& \multicolumn{3}{c}{\textbf{NoTag}} \\
\cmidrule(lr){2-4} \cmidrule(lr){5-7} \cmidrule(lr){8-10} \cmidrule(lr){11-13} \cmidrule(lr){14-16} \cmidrule(lr){17-19}
\cmidrule(lr){20-22} \cmidrule(lr){23-25}
\textbf{Model} & {T.} & {I.} & {G.} & {T.} & {I.} & {G.} & {T.} & {I.} & {G.} & {T.} & {I.} & {G.} &
{T.} & {I.} & {G.} & {T.} & {I.} & {G.} & {T.} & {I.} & {G.} & {T.} & {I.} & {G.} \\
\midrule
CLIP & 31.3 & 11.3 & 9.0 & \textbf{76.7} & 40.0 & 36.7 & \textbf{25.0} & 8.9 & 5.4 & 15.8 & 0.0 & 0.0 & \textbf{36.0} & 14.0 & 10.0 & 30.4 & \textbf{15.2} & \textbf{15.2} & 24.4 & 6.4 & \textbf{3.8} & 32.0 & 11.6 & 9.3 \\
SigLIP & 32.8 & 12.8 & 10.3\\
\cdashline{1-27}[0.5pt/1pt]
DAC$_{LLM}$ & 22.5 & 10.3 & 4.8 & 50.0 & 26.7 & 20.0 & 16.1 & 10.7 & 3.6 & 10.5 & 10.5 & 7.9 & 20.0 & 10.0 & 2.0 & 19.6 & 10.9 & 4.3 & 19.2 & \textbf{10.3} & \textbf{3.8} & 23.8 & 9.9 & 3.5 \\
DCI$_{P1}$ & 20.8 & 10.3 & 5.3 & 53.3 & 26.7 & 23.3 & 23.2 & 5.4 & 5.4 & 21.1 & 2.6 & 2.6 & 18.0 & 4.0 & 2.0 & 15.2 & 10.9 & 4.3 & 19.2 & \textbf{10.3} & \textbf{3.8} & 18.0 & 11.0 & 4.7 \\
CE-CLIP & 19.5 & 12.0 & 5.3 & 36.7 & 33.3 & 13.3 & 12.5 & \textbf{16.1} & \textbf{8.9} & 5.3 & 7.9 & 0.0 & 18.0 & 16.0 & 6.0 & 23.9 & 10.9 & 6.5 & 17.9 & 7.7 & 2.6 & 20.3 & 11.0 & 4.7 \\
NegCLIP & 30.3 & 11.0 & 8.0 & 66.7 & 30.0 & 26.7 & 17.9 & 8.9 & 3.6 & 10.5 & 2.6 & 2.6 & \textbf{36.0} & 10.0 & 8.0 & 28.3 & 4.3 & 4.3 & 21.8 & 9.0 & 5.1 & 35.5 & 12.2 & 8.7 \\
\cdashline{1-27}[0.5pt/1pt]
LongCLIP-B & 28.5 & 8.8 & 7.3 & 66.7 & 40.0 & \textbf{40.0} & \textbf{25.0} & 5.4 & 5.4 & \textbf{28.9} & 2.6 & 2.6 & 28.0 & 14.0 & 12.0 & 32.6 & 8.7 & 8.7 & 25.6 & 0.0 & 0.0 & 26.7 & 8.7 & 5.8 \\
DreamLIP & 28.2 & \textbf{15.0} & \textbf{10.8} & 46.7 & \textbf{43.3} & 33.3 & 23.2 & 10.7 & 7.1 & 23.7 & \textbf{13.2} & \textbf{10.5} & 24.0 & \textbf{18.0} & \textbf{14.0} & \textbf{34.8} & 13.0 & 8.7 & \textbf{28.2} & 9.0 & 5.1 & 26.7 & \textbf{17.4} & \textbf{13.4} \\
LSS & \textbf{33.3} & 11.8 & 7.5 & 66.7 & 26.7 & 26.7 & \textbf{25.0} & 8.9 & 3.6 & 18.4 & 0.0 & 0.0 & 32.0 & 8.0 & 8.0 & 32.6 & \textbf{15.2} & 13.0 & \textbf{28.2} & 7.7 & 2.6 & \textbf{36.0} & 14.0 & 8.1 \\
\bottomrule
\end{tabular}
}
\caption{
\textbf{Full Winoground Performance and Tag Breakdown.} The best-performing model on grouped results varies across tags. LSS demonstrates consistently strong textual understanding, achieving the highest overall score and leading in several categories, including complex reasoning. These results suggest that training with long captions helps improve performance on this benchmark.}
\label{tab:winoground-full}
\end{table*}

\section{SigLIP Results}
\label{app:siglip-results}
To complement our main results and validate that our findings generalize beyond CLIP-based architectures, we report results for SigLIP~\cite{zhai2023sigmoid}, a vision–language model trained with a sigmoid loss on the large-scale WebLI dataset~\cite{chen2022pali}. As shown in Table~\ref{tab:architecture-variation}, SigLIP demonstrates stronger generalization on classification and short retrieval tasks compared to CLIP-based models, DreamLIP and LSS (ViT-B/16).

Despite the absence of explicit long-caption supervision, SigLIP achieves performance comparable to DreamLIP and LSS/16 on long-caption retrieval benchmarks and performs strongly on Winoground and SC++. This consistency across tasks supports our central conclusion that long-caption understanding and compositional reasoning are mutually reinforcing capabilities. Importantly, SigLIP’s results, achieved with a different objective function, training data, and architecture, demonstrate that these relationships are not artifacts of CLIP’s design but rather may reflect general principles of vision–language learning.

\section{Full Winoground Results}
\label{app:winoground-groups}
We report the complete Winoground results, including separate scores for Text, Image, and Group. In addition, we group results by the tags proposed by \citet{diwan2022winoground}, which break the benchmark into finer-grained subcategories and allow for deeper analysis (see Table \ref{tab:winoground-full}). 

Across the full dataset (Overall columns), LSS achieves the highest Text score and competitive performance on Image and Group. When grouped by tags, LSS consistently shows strong Text performance, often ranking first or close to the best-performing model. Interestingly, vanilla CLIP achieves strong results on several individual groups, highlighting that Winoground remains challenging for current representational learning approaches. These results reinforce that, while progress is being made, compositional reasoning in VLMs is still far from solved.

\begin{table*}
\centering
\resizebox{.9\textwidth}{!}{%
\begin{tabular}{@{} l
S[table-format=7.0] %
S[table-format=7.0] %
c %
c %
S[table-format=2.2] %
S[table-format=3.2] %
c %
@{}}
\toprule
\textbf{Dataset} & \textbf{Images} & \textbf{Captions} & \textbf{Avg. Length} & \textbf{Word Range} & \textbf{Vocab. Covered (\%)} & \textbf{Yngve} & \textbf{Source} \\
\midrule
sDCI$_{\text{Train}}$ & 7599 & 82785 & 40 $\pm$ 12 & 4 - 70 & 29.29 & 94.07 & Mixed\\
DOCCI$_{\text{Train}}$ & 14647 & 14647 & 122 $\pm$ 45 & 28 - 518 & 26.96 & 74.55 & Human\\
LN & 489000 & 489000 & 30 $\pm$ 17 & 1 - 226 & 24.34 & 61.70 & Human\\
ShareGPT4v & 1200000 & 1200000 & 144 $\pm$ 39 & 19 - 507 & 87.72 & 45.70 & Synthetic\\
\cdashline{1-8}[0.5pt/1pt] %
COCO$_{\text{Test}}$     & 5000    & 24855     & 10 $\pm$ 3 & 6 - 43 & 14.75 & 24.53 & Human \\
Flickr30k$_{\text{Test}}$ & 1000      & 4999      & 12 $\pm$ 5 & 2 - 68 & 8.99 & 34.64 & Human \\
Urban1k & 1000      & 1000    & 107 $\pm$ 10 & 74 - 179 & 10.75 & 69.27 & Synthetic \\
IiW$_{\text{Test}}$ & 400 & 400 & 217 $\pm$ 83 & 45 - 480 & 11.53 & 119.22 & Mixed\\
sDCI$_{\text{Test}}$ & 206    & 2236 & 41 $\pm$ 12 & 11 - 67 & 7.67 & 95.64 & Mixed\\
DOCCI$_{\text{Test}}$ & 5200 & 5200 & 123 $\pm$ 46 & 28 - 518 & 19.29 & 73.95 & Human\\
\bottomrule
\end{tabular}
}
\caption{\textbf{Dataset Statistics for Training and Evaluation}.
Datasets above the dashed line are used for training, while those below are used for testing. The `Word Range' column indicates the words in the shortest and longest caption in the dataset and the `Vocabulary Covered (\%)' column indicates the percentage of tokens found in the respective datasets that are also present in the tokenizer training vocabulary of CLIP. The `Yngve' metric quantifies syntactic complexity, with higher values indicating deeper left-branching structures~\cite{yngve1960model}. Many datasets exceed CLIP’s 77-token input limit, especially those designed for long-caption understanding.
}
\label{tab:dataset-stats}
\end{table*}

\section{Compositionality and Long-Caption Retrieval Results for LLS Models.}
\label{app:LLS-results}
We provide a detailed breakdown of the performance for our LSS model variants. Table \ref{tab:LSS-retrieval} presents the complete results for LSS models trained on different datasets of image and long-caption pairs. The reported metrics cover both long-caption retrieval and compositionality benchmarks, offering a comprehensive overview that allows for a direct comparison of how different training data affects final model capabilities.

\begin{table*}[t]
\centering
\resizebox{\textwidth}{!}{%
\begin{tabular}{@{} l
    *{7}{S[table-format=2.1]} %
    c %
    *{9}{S[table-format=2.1]} %
    @{}}
\toprule
& \multicolumn{7}{c}{\textbf{Compositional Reasoning}}
&
& \multicolumn{9}{c}{\textbf{Long-Caption Retrieval}} \\
\cmidrule(lr){2-8} \cmidrule(lr){10-18}
& & \multicolumn{6}{c}{\textbf{SugarCrepe++}} &
& \multicolumn{2}{c}{\textbf{Urban1k}}
& \multicolumn{2}{c}{\textbf{sDCI}}
& \multicolumn{2}{c}{\textbf{DOCCI}}
& \multicolumn{2}{c}{\textbf{IiW}} & \\ %
\cmidrule(lr){3-8} \cmidrule(lr){10-11} \cmidrule(lr){12-13} \cmidrule(lr){14-15} \cmidrule(lr){16-17}
\textbf{Model} & {WG} & {SA} & {RR} & {RO} & {RA} & {SO} & {Avg.} %
& \multicolumn{1}{c}{} %
& {I2T} & {T2I} & {I2T} & {T2I} & {I2T} & {T2I} & {I2T} & {T2I} & {Avg.} \\ %
\midrule
LSS$_{sDCI}$         & \textbf{18.5} & 47.0 & 51.0 & 90.5 & 69.4 & 29.2 & 57.4 & \multirow{4}{*}{\boldmath{$\leftarrow$}} & 61.2 & 58.0 & 90.3 & \textbf{78.7} & 54.7 & 54.3 & 87.5 & 88.0 & 71.6 \\
LSS$_{DOCCI}$        & 18.0 & 51.2 & \textbf{56.3} & 90.1 & 74.4 & 32.4 & 60.9 & & 67.2 & 62.2 & 89.3 & 76.6 & \textbf{90.6} & \textbf{91.1} & 93.3 & 91.5 & 82.7 \\
LSS$_{LN}$           & 12.9 & 53.5 & 52.9 & \textbf{93.1} & 73.6 & 34.7 & 61.6 & & 57.8 & 58.3 & 88.8 & 76.3 & 54.1 & 55.0 & 88.3 & 87.0 & 70.7\\
LSS$_{ShareGPT4V}$   & 17.5 & 52.2 & 53.4 & 91.3 & \textbf{74.9} & \textbf{36.5} & \textbf{61.7} & & 75.4 & 74.1 & \textbf{91.7} & 75.1 & 64.5 & 63.0 & \textbf{94.0} & 92.0 & 78.7 \\
\bottomrule
\end{tabular}%
}
\caption{
\textbf{Compositional (left) and Long-Caption Retrieval (right) Performance Across Models.}}
\label{tab:LSS-retrieval}
\end{table*}

\end{document}